\documentclass[12pt]{article}
\usepackage{amsmath}
\usepackage{graphicx,psfrag,epsf}
\usepackage{enumerate}
\usepackage{natbib}
\usepackage{amsfonts}
\usepackage{xcolor}
\usepackage{url} % not crucial - just used below for the URL 

\usepackage{comment}
\usepackage{amsthm}
\newtheorem{theorem}{Theorem}

\theoremstyle{definition}

\theoremstyle{remark}

\usepackage{listings}

\usepackage{algorithm}
\usepackage{algpseudocode}
\usepackage{amsmath}

\usepackage[utf8]{inputenc} % allow utf-8 input
\usepackage[T1]{fontenc}    % use 8-bit T1 fonts
\usepackage{hyperref}       % hyperlinks
\usepackage{url}            % simple URL typesetting
\usepackage{booktabs}       % professional-quality tables
\usepackage{amsfonts}       % blackboard math symbols
\usepackage{nicefrac}       % compact symbols for 1/2, etc.
\usepackage{microtype}      % microtypography
\usepackage{xcolor}         % colors
\usepackage{graphicx}
\usepackage{gensymb}
\usepackage{caption}
\newcommand{\blind}{1}

\def\spacingset#1{\renewcommand{\baselinestretch}%
{#1}\small\normalsize} \spacingset{1}

\begin{document}

\if1\blind
{
  
  \title{\bf Uncertainty-Aware Out-of-Distribution Detection with Gaussian Processes}
  {\author{Yang Chen\hspace{.2cm}\\
    School of Computer Science, Carnegie Mellon University\\
    and\\
    Chih-Li Sung\\
    Department of Statistics and Probability, Michigan State University\\
    and\\
    Arpan Kusari \\
    University of Michigan Transportation Research Institute\\
    and\\
    Xiaoyang Song\\
    Department of Industrial and Operations Engineering, University of Michigan\\
    and\\
    Wenbo Sun \\
    University of Michigan Transportation Research Institute}}
        \date{\vspace{-1ex}}
  %\maketitle
} \fi

\if0\blind
{
  \bigskip
  \bigskip
  \bigskip
  \begin{center}
    \title{\LARGE\bf Uncertainty-Aware Out-of-Distribution Detection with Gaussian Processes}
\end{center}
  \medskip
} \fi

\bigskip

% The \author macro works with any number of authors. There are two commands
% used to separate the names and addresses of multiple authors: \And and \AND.
%
% Using \And between authors leaves it to LaTeX to determine where to break the
% lines. Using \AND forces a line break at that point. So, if LaTeX puts 3 of 4
% authors names on the first line, and the last on the second line, try using
% \AND instead of \And before the third author name.

\newcommand{\wenbo}[1]{\textcolor{magenta}{#1}}
\newcommand{\arpan}[1]{\textcolor{green}{#1}}
\newcommand{\chihli}[1]{\textcolor{blue}{#1}}
\newcommand{\yang}[1]{\textcolor{cyan}{#1}}
\maketitle

\begin{abstract}
Deep neural networks (DNNs) are often constructed under the closed-world assumption, which may fail to generalize to the out-of-distribution (OOD) data. This leads to DNNs producing overconfident wrong predictions and can result in disastrous consequences in safety-critical applications. Existing OOD detection methods mainly rely on curating a set of OOD data for model training or hyper-parameter tuning to distinguish OOD data from training data (also known as in-distribution data or InD data). However, OOD samples are not always available during the training phase in real-world applications, hindering the OOD detection accuracy. To overcome this limitation, we propose a Gaussian-process-based OOD detection method to establish a decision boundary based on InD data only. The basic idea is to perform uncertainty quantification of the unconstrained softmax scores of a DNN via a multi-class Gaussian process (GP), and then define a score function to separate InD and potential OOD data based on their fundamental differences in the posterior predictive distribution from the GP. Two case studies on conventional image classification datasets and real-world image datasets are conducted to demonstrate that the proposed method outperforms the state-of-the-art OOD detection methods when OOD samples are not observed in the training phase.

\end{abstract}

\section{Introduction}
\label{sec:intro}
In recent years, deep neural networks (DNNs) have been widely used in various fields due to their exceptional performance in classification tasks. However, the performance of DNNs can deteriorate significantly when tested on out-of-distribution (OOD) data, which refers to data comes from a different distribution than the training (In-distribution or InD) data distribution. This happens because DNNs tend to classify test data to a known InD data class with high certainty, no matter if it comes from an InD image group or not. The presence of OOD data can lead to poor reliability of DNNs. However, in real-life scenarios, OOD data is common, such as in autonomous driving \citep{henriksson2023out} and medical diagnosis \citep{hong2024out}. Thus, there is a critical need for a DNN to detect OOD data before making predictions. 

The necessity of detecting OOD data in DNNs was first discussed by \citet{nguyen2015deep} and has gained increasing attention since then. The current literature on OOD detection have three main categories. The first category is score-based OOD detection, which mainly utilizes pre-defined score functions to project InD and OOD data to different extrema in the score continuum. For example, \citet{lee2018simple} used a confidence threshold on the last layer of the DNN to identify OOD data. \citet{liang2018enhancing} introduced uncertainty to the penultimate layer of the DMM and added perturbations to the input data using a gradient-based method. \citet{liu2020energy} developed the energy score of samples to separate OOD data from InD data. \citet{van2020uncertainty} leveraged RBF networks to define a score function that quantifies the prediction uncertainty. With a well-trained score function, these methods can perform OOD detection on pre-trained DNNs by simply adding an additional module without the need for re-training.

The second category is retraining-based OOD detection, which enables OOD detection through retraining DNNs. \citet{hendrycks2019using} added an auxiliary network to the original model and trained it using self-supervised learning to improve the model's robustness and uncertainty. \citet{chen2021enhancing} modified the empirical loss function of the DNN to assign a high entropy on each OOD sample, and hence identified OOD data. These methods often achieve higher OOD detection accuracies compared to the score-based methods. However, methods in the first two categories rely on hyper-parameters which are required to be tuned using OOD data. In real-world scenarios, OOD data are not always observable, making these methods fail to generalize to detect potential OOD samples. 

The third category of OOD detection methods emerged with the development of generative models. These methods regularize the training loss using virtual OOD data generated based on InD data distribution. For instance, \citet{du2022vos} sampled virtual OOD data from the tail of Gaussian distribution which InD data was assumed to follow. \citet{tao2023non} generated virtual OOD data around the boundary InD samples identified from the nearest neighbor method and then established a decision boundary between InD and OOD data. However, these generative approaches either result in poor generalization performance or require hyper-parameter tuning for the temperature scaling and input perturbation magnitudes. Theoretical guarantees are also not provided in this category of methods.

In this paper, we aim to develop an OOD detection method that does not require OOD data for model training or hyper-parameter tuning. The proposed method is motivated by Statistical Quality Control \citep{montgomery2020introduction}, that is, to quantify a specific monitoring feature of InD data and generate alarms for incoming data whose features lie out of the control limit. In particular, we leverage the Gaussian process (GP) \citep{rasmussen2006gaussian,gramacy2020surrogates} to produce high prediction uncertainty at OOD data due to the fact that GP shows poor extrapolation performance. A score function is then defined to capture the discrepancy in predictive distribution at InD and OOD samples. The contributions of the paper are three fold: (i) The OOD detection model training does not rely on any OOD data; (ii) The proposed method is compatible with most existing DNN architectures; (iii) Statistical theories are developed for OOD detection performance. 

The remainder of the paper is organized as follows. In Section~\ref{s:notation}, we will describe the notation, formulation, and objective. In Section~\ref{s:method}, the proposed method is elaborated for OOD detection. Section~\ref{s:case} demonstrates the effectiveness of the method by using two different setups in image classification tasks. The paper then ends with conclusions and discussions in Section~\ref{s:conclusion}.

\section{Notation, Formulation, and Objective}\label{s:notation}
The paper focuses on OOD detection in classification tasks. Let $\mathcal{X}_\text{InD}$ and $\mathcal{X}_\text{OOD}$ represent the spaces of InD and OOD data, respectively. Let $\mathcal{Y}_\text{InD}$ denote the set of labels for the InD data, where $\left|\mathcal{Y}_\text{InD}\right|=K$ corresponds to the number of class labels. In conventional DNN classification, a neural network $\mathbf{f}$ maps a $d$-dimensional input $\mathbf{x}\in\mathcal{X}_\text{InD}\cup\mathcal{X}_\text{OOD}:=\mathcal{X}\subseteq\mathbb{R}^d$ to a vector of length  $K$, $\mathbf{f}(\mathbf{x}):=(f_1(\mathbf{x}),\ldots,f_K(\mathbf{x}))$, and predicts the class label of $\mathbf{x}$ as 
$$k^*=\text{arg}\,\max_{k}{f}_k(\mathbf{x}).$$ 
Typically, the activation function of the last layer is chosen as the \textit{Softmax} function, so that the last layer reflects the probability of $\mathbf{x}$ belonging to the $k$-th class. In this work, we use the unconstrained version of the Softmax function for $\mathbf{f}(\mathbf{x})$ to avoid the potential non-smoothness introduced by normalization, which could adversely affect subsequent modeling steps.

In OOD detection, for an input $\mathbf{x}\in\mathcal{X}_\text{OOD}$, a conventional DNN tends to classify $\mathbf{x}$ into a known class with a high confidence, i.e., producing a $\mathbf{f}(\mathbf{x})$ close to a one-hot vector whose $k$-th element is one while others are zeros. Consequently, $\mathbf{x}$ may be mistakenly treated as an InD sample from class $k$.  To address this issue, it is essential to introduce an OOD detection function $g$ to identify OOD samples before making classification decisions with $\mathbf{f}$, where $g$ is defined as:
\begin{equation}
g(\mathbf{x})=\left\{\begin{array}{l}
1, \text { if } \mathbf{x} \in \mathcal{X}_{\text{OOD}} \\
0, \text { if } \mathbf{x} \in \mathcal{X}_\text{InD}
\end{array}\right.\nonumber
\end{equation}
The definition of $g$ has been extensively explored in the machine learning literature, typically relying on a score function that separates InD and OOD samples. For instance, in \citet{liang2018enhancing}, the score function is based on the calibrated Softmax score: 
\begin{equation}
    g_T(\mathbf{x})=\mathbb{I}\left\{\max_k\,\exp \left(f_k(\mathbf{x}+\mathbf{\delta}) / T\right)> \gamma\right\},
    \label{eq:odin}
\end{equation}
where $T$, $\delta$ and $\gamma$ denote the temperature scaling parameter, disturbance magnitude, and detection threshold, respectively. These hyper-parameters are tuned to optimize the separation between $\mathcal{X}_\text{InD}$ and $\mathcal{X}_\text{OOD}$. 

Existing methods for OOD detection often require a substantial number of OOD samples to tune hyper-parameters such as $T$, $\boldsymbol{\delta}$, and $\gamma$ in \eqref{eq:odin}. However, in practical scenarios, collecting a sufficient number of OOD samples is not always feasible. To address this limitation, we propose a score function $s(\mathbf{x}; \theta)$ parameterized by $\theta$, and define the OOD detection function as: $$g(\mathbf{x})=\mathbb{I}\left\{s(\mathbf{x};\theta)>\gamma\right\},$$ where both $\theta$ and $\gamma$ are trained \textit{exclusively with InD samples}. In the next section, we elaborate on the definition and estimation of $s$, $\theta$, and $\gamma$.

\section{OOD Detection via Gaussian Processes}\label{s:method}
\begin{figure}
    \centering
    \includegraphics[width=0.95\textwidth]{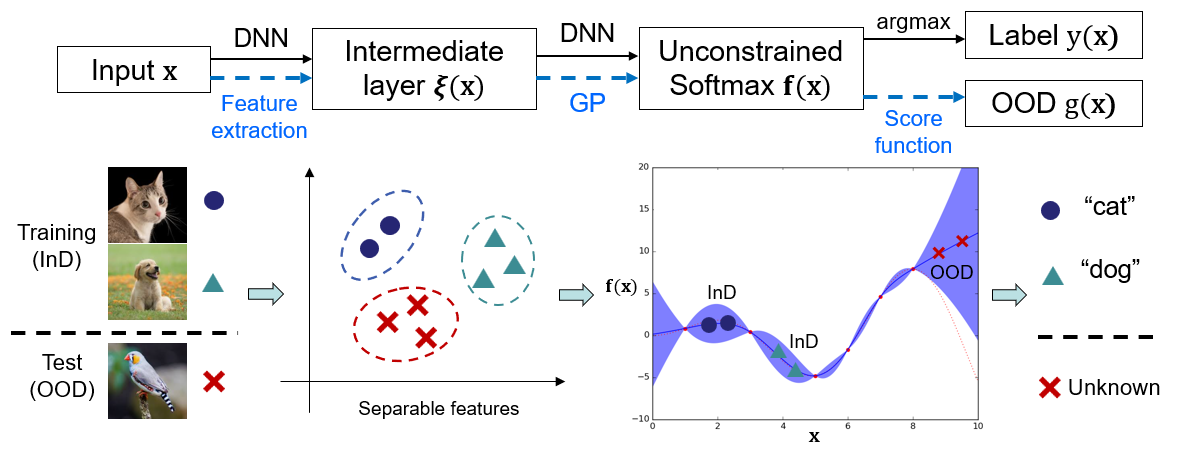}
    \caption{Flowchart of uncertainty-aware OOD detection via a multi-class GP model.}
    \label{fig:flowchart}
\end{figure}

The score function $s$ is constructed to leverage the prediction uncertainty of DNNs.  Given that DNNs typically yield deterministic predictions, we employ a Gaussian Process (GP) model to emulate the target DNN and quantify the associated prediction uncertainty. Intuitively, since the GP emulator is trained exclusively on InD data, it exhibits higher prediction uncertainty at $\mathcal{X}_\text{OOD}$  compared to $\mathcal{X}_\text{InD}$. This behavior stems from the well-known property of GPs, where poor extrapolation leads to higher prediction uncertainty and bias. Motivated by this property, the proposed OOD detection method consists of two key modules: (i) a multi-class GP to quantify the prediction uncertainty of the DNN (introduced in Section \ref{sec:MultiGP-method}), and (ii) a score function to separate OOD data from InD data without requiring access to OOD samples (introduced in Section \ref{ss:score}). The proposed method is summarized in the flowchart provided in Figure~\ref{fig:flowchart}. 

\subsection{Multi-Class GP Modeling for Uncertainty Quantification}\label{sec:MultiGP-method}
The first step is to quantify the uncertainty of the DNN predictions $\mathbf{f}$. Since DNNs often handle high-dimensional inputs, such as images or tensors, directly constructing a GP with input $\mathbf{x}$ is computationally challenging \citep{stork2020open}. Instead, we use an intermediate layer of the DNN as the input to the GP, denoted by $\boldsymbol\xi(\mathbf{x})\in\mathbb{R}^p$,  which reduces the dimensionality of the input. The GP aims to predict $\mathbf{f}(\mathbf{x})$ using the input $\boldsymbol\xi(\mathbf{x})$. It is desirable that $\left\{\boldsymbol\xi(\mathbf{x})|\mathbf{x}\in\mathcal{X}_\text{InD}\right\}$ and $\left\{\boldsymbol\xi(\mathbf{x})|\mathbf{x}\in\mathcal{X}_\text{OOD}\right\}$ are separable. This argument will be revisited in Subsection~\ref{ss:intermediate}. %This argument leads to a key assumption on the minimum distance between InD and OOD data points under a specific distance metric, which we resivit in Subsection~\ref{ss:intermediate}.

To model the predictions $\boldsymbol{f}(\mathbf{x})$ using $\boldsymbol{\xi}(\mathbf{x})$, we construct a multi-class GP. Since the response $\boldsymbol{f}(\mathbf{x})$ is a $K$-dimensional vector, we fit $K$ separate GP models, one for each class. Let $\mathcal{X}^k_{\text{InD}}$ denote the InD data belonging to class $k$. Conventional GP models often assume a stationary covariance function that depends solely on relative distances in the input space. However, when $\mathcal{X}_\text{InD}$ is comprised of data from multiple classes, the estimation of GP parameters in the covariance function for $f_k(\mathbf{x})$ can be affected by the data from $\mathcal{X}_{\text{InD}}^j$ for $j\neq k$, which often leads to poor predictions of $f_k(\mathbf{x})$. To address this, we model the non-stationary covariance structure using separate GP models for each class. That is, for each class $k$, a GP model is trained to map $\boldsymbol{\xi}_k (\mathbf{x})$ to $f_k(\mathbf{x})$ using the data from $\mathcal{X}_\text{InD}^k$. 

Recall that $f_k(\mathbf{x})$ is the unconstrained Softmax score of label $k$. Assume the unconstrained Softmax score $f_k$  follows a zero-mean GP prior:
\begin{equation}
    f_k(\mathbf{x}) \sim \mathcal{GP}\left(0,\tau^2_k\Phi_{k}(\boldsymbol{\xi}(\mathbf{x}),\boldsymbol{\xi}(\mathbf{x}'))\right)\quad\text{for}\quad k=1,\ldots,K,\nonumber
\end{equation}
where $\tau^2_k>0$ is the scale hyper-parameter and $\Phi_k\left(\cdot,\cdot\right)\in\mathbb{R}^p\times\mathbb{R}^p\rightarrow \mathbb{R} $ is a positive definite kernel. Throughout this paper, we use the squared exponential kernel: 
\begin{align}
\Phi_k(\boldsymbol{\xi}(\mathbf{x}), \boldsymbol{\xi}(\mathbf{x}'))=\exp \left( -\sum_{j=1}^p \frac{\left(\xi_j(\mathbf{x})-\xi_j(\mathbf{x}')\right)^2}{\theta_{k,j}}\right),
    \label{eq:kernel function}
\end{align}
where $\theta_{k,j}$ is the lengthscale hyper-parameter. 

Given the InD data with label $k$, denoted by $X_{\rm{InD}}^k\subset \mathcal{X}_{\text{InD}}^k$, of size $n_k$, with corresponding unconstrained Softmax scores  $\mathbf{z}_{\rm{InD}}^k:=\{f_k(\mathbf{x}):\mathbf{x}\in X_{\rm{InD}}^k\}\in\mathbb{R}^{n_k\times 1}$ , the predictive distribution of $f_k(\mathbf{x})$ at a new input is:$$f_k(\mathbf{x})|\mathbf{z}_{\rm{InD}}^k\sim\mathcal{N}(\mu_k(\mathbf{x}), \sigma^2_k(\mathbf{x})),$$ with the mean and variance:
\begin{equation*}
     \mu_k(\mathbf{x}) = \Phi_k(\boldsymbol{\xi}(\mathbf{x}),\boldsymbol{\xi}(X_{\rm{InD}}^k)) \Phi_k(\boldsymbol{\xi}(X_{\rm{InD}}^k),\boldsymbol{\xi}(X_{\rm{InD}}^k))^{-1} \mathbf{z}_{\rm{InD}}^k,
 \end{equation*}
and
 \begin{equation*}
     \sigma_k^2(\mathbf{x}) = \tau^2_k\left[1 - \Phi_k(\boldsymbol{\xi}(\mathbf{x}),\boldsymbol{\xi}(X_{\rm{InD}}^k))\Phi_k(\boldsymbol{\xi}(X_{\rm{InD}}^k),\boldsymbol{\xi}(X_{\rm{InD}}^k))^{-1} \Phi_k(\boldsymbol{\xi}(X_{\rm{InD}}^k),\boldsymbol{\xi}(\mathbf{x}))\right],
 \end{equation*}
where $\Phi_k(\boldsymbol{\xi}(\mathbf{x}),\boldsymbol{\xi}(X_{\rm{InD}}^k))$ is defined as the collection of kernel evaluations between $\boldsymbol{\xi}(\mathbf{x})$ and all points in $X_{\rm{InD}}^k$:
$$\Phi_k(\boldsymbol{\xi}(\mathbf{x}),\boldsymbol{\xi}(X_{\rm{InD}}^k))=\{\Phi_k(\boldsymbol{\xi}(\mathbf{x}),\boldsymbol{\xi}(\mathbf{x}')):\mathbf{x}'\in X_{\rm{InD}}^k\}\in\mathbb{R}^{1\times n_k},$$ which is equivalent to $\Phi_k(\boldsymbol{\xi}(X_{\rm{InD}}^k),\boldsymbol{\xi}(\mathbf{x}))^\top$. Similarly, $\Phi_k(\boldsymbol{\xi}(X_{\rm{InD}}^k),\boldsymbol{\xi}(X_{\rm{InD}}^k))$ denotes the kernel matrix for all pairs of points in $X_{\rm{InD}}^k$:
$$\Phi_k(\boldsymbol{\xi}(X_{\rm{InD}}^k),\boldsymbol{\xi}(X_{\rm{InD}}^k))=\{\Phi_k(\mathbf{x},\mathbf{x}'):\mathbf{x},\mathbf{x}'\in X_{\rm{InD}}^k\}\in\mathbb{R}^{n_k\times n_k}.$$
These definitions will be similarly used  throughout the paper.

The predictive distribution provides both the mean prediction and uncertainty quantification of $f_k(\mathbf{x})$. Based on this, a score function is developed to detect OOD samples using the GP predictions. The parameters associated with the kernel function \eqref{eq:kernel function}, including $\tau^2_k$ and $\theta_{k,j}$, are estimated via maximum likelihood estimation (MLE). Specifically, for fixed $\theta_{k,j}$, the MLE of $\tau^2_k$ is given by \citep{gramacy2020surrogates}:
\begin{equation}\label{eq:MLEtau2}
\hat{\tau}^2_k=\frac{1}{n_k}(\mathbf{z}_{\rm{InD}}^k)^\top\Phi_k(\boldsymbol{\xi}(X_{\rm{InD}}^k),\boldsymbol{\xi}(X_{\rm{InD}}^k))^{-1}_k\mathbf{z}_{\rm{InD}}^k.
\end{equation}
Substituting $\hat{\tau}_k^2$ back into the log-likelihood yields a \textit{profile} log-likelihood that depends only on the remaining parameter $\theta_{k,j}$:
\[
\ell(\{\theta_{k,j}\}^d_{j=1}) = c - \frac{n_k}{2}\log\left((\mathbf{z}_{\rm{InD}}^k)^\top \Phi_k(\boldsymbol{\xi}(X_{\rm{InD}}^k),\boldsymbol{\xi}(X_{\rm{InD}}^k))^{-1}\mathbf{z}_{\rm{InD}}^k\right) - \frac{1}{2}\log|\Phi_k(\boldsymbol{\xi}(X_{\rm{InD}}^k),\boldsymbol{\xi}(X_{\rm{InD}}^k))|,
\]
where $c$ is a constant. The profile log-likelihood can then be maximized using optimization algorithms, such as the quasi-Newton method proposed by \cite{byrd1995limited}.

\subsection{GP-Based Detection Score Function}\label{ss:score}
The detection score $s(\mathbf{x}')$ is formulated based on the predictive distribution of the multi-class GP.  For two samples $(\mathbf{x}, \mathbf{x}')$ belonging to the same class $k$, it is reasonable to expect their predictive distributions for the unconstrained Softmax scores to be similar. We choose to use the Kullback-Leibler (KL) divergence as the distance metric between the two predictive distributions of $\eta_k(\mathbf{x})$ and $\eta_k(\mathbf{x}')$:
\begin{eqnarray}
    d\left(\mathbf{x}',\mathbf{x}\right)&=&\text{KL}\left(\mathcal{N}\left(\mu_k(\mathbf{x}'),\sigma^2_k(\mathbf{x}')\right),\mathcal{N}\left(\mu_k(\mathbf{x}),\sigma^2_k(\mathbf{x})\right)\right)\nonumber\\
    &=& \log \frac{\sigma^2_k(\mathbf{x})}{\sigma^2_k(\mathbf{x}')} + \frac{\sigma^2_k(\mathbf{x}') + (\mu_k(\mathbf{x}') - \mu_k(\mathbf{x}))^2}{2 \sigma_k(\mathbf{x})^2} - \frac{1}{2}.
    \label{eq:KL}
\end{eqnarray}

\begin{figure}[h]
    \centering
    \includegraphics[width=0.9\linewidth]{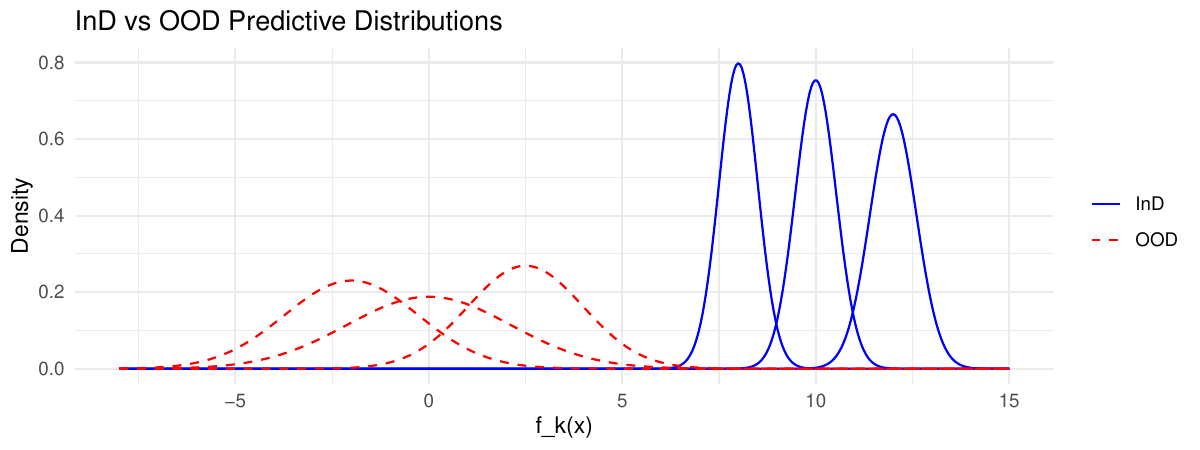}
    \caption{Illustration of InD (in-distribution) predictive distributions (blue, solid lines) with smaller variances and larger means, compared to OOD (out-of-distribution) predictive distributions (red, dashed lines) with larger variances and smaller means. The x-axis represents the predicted values $f_k(\mathbf{x})|\mathbf{z}_{\rm{InD}}^k$, and the y-axis represents the density of the predictive distributions.}
    \label{fig:KLdemo}
\end{figure}

Figure \ref{fig:KLdemo} provides an illustrative example of the KL divergence between InD and OOD samples. Only three InD distributions and three OOD distributions are displayed for simplicity. The InD predictive distributions (blue, solid lines) typically have means around 10, corresponding to the range of the unconstrained Softmax scores $\mathbf{z}^k_{\rm{InD}}$ observed during training. These distributions also have small variances, reflecting the model's confidence in its predictions for in-distribution samples. In contrast, the OOD predictive distributions (red, dashed lines) show means closer to zero. This occurs because, as test samples move farther from the training data, the predictions degenerate towards zero, which is the prior mean \citep{santner2018design}. Additionally, OOD distributions display larger variances, representing the increased uncertainty when extrapolating beyond the training data. This combination of characteristics—small variance and higher means for InD, and large variance with lower means for OOD, results in significant KL divergence between $\mathbf{x}\in\mathcal{X}^k_{\rm{InD}}$ and $\mathbf{x}'\in\mathcal{X}^k_{\rm{OOD}}$, facilitating robust differentiation between InD and OOD samples.

Subsequently, for  a test sample $\mathbf{x'}$, whether originating from InD or OOD, we ascertain its predicted classification via $\boldsymbol{f}(\mathbf{x}')$ derived from the DNN. Suppose that $\mathbf{x'}$ is predicted to belong to class $k$, i.e., $k=\text{arg}\,\max_{l=1,\ldots,K}{f}_{l}(\mathbf{x}')$, then the detection score for $\mathbf{x'}$ is  defined as the average KL divergence between $\mathbf{x'}$ and the InD samples $X_{\rm{InD}}^k$:
\begin{equation}
s_k(\mathbf{x'}) = \frac{1}{n_k}\sum_{\mathbf{x}\in X_{\rm{InD}}^k}d(\mathbf{x}',\mathbf{x}).\nonumber
\end{equation}
The detection function $g$ is then:
\begin{equation}
g(\mathbf{x'})=\left\{\begin{array}{l}
1, \text { if } s_k(\mathbf{x'}) > \gamma_k\quad\text{(OOD)}\\
0, \text { if } s_k(\mathbf{x'}) \leq \gamma_k\quad\text{(InD)},\nonumber
\end{array}\right.
\end{equation}
This implies that if a test sample $\mathbf{x'}$ is farther from $X_{\rm{InD}}^k$ than the critical threshold $\gamma_k$, it is likely to be classified as an OOD sample.

The critical value $\gamma_k$ is determined \textit{solely using the InD samples}. Specifically, for two InD data points $\mathbf{x}$ and $\mathbf{x}'$ belonging to class $k$, a probabilistic upper bound on $d\left(\mathbf{x}, \mathbf{x}'\right)$ can be derived from the $1-\alpha$ quantile, $q_\alpha$, of the distances computed within the training InD samples. This is expressed as
\begin{equation} \gamma_k = q_\alpha\left(s_k(\mathbf{x'}) \text{ for all } \mathbf{x}' \in X_{\text{InD}}^k\right).\nonumber 
\end{equation}
However, using the same set of training InD samples for both fitting the GPs and determining $\gamma_k$ can lead to overly conservative OOD detection due to the interpolation properties of GPs \citep{santner2018design}. Specifically, at the training points, the GP predictive variance $\sigma_k^2(\mathbf{x})$ becomes zero, and the predictive mean $\mu_k^2(\mathbf{x})$ equals $f_k(\mathbf{x})$. To address this issue, the training InD samples are divided into two subsets (e.g., 80\% and 20\%). The larger subset is used to train the GPs, while the smaller subset is reserved for OOD detection (i.e., constructing $g(\mathbf{x}')$), which we call \textit{validation set}. This approach ensures a more robust detection function $g(\mathbf{x}')$, as outlined in Algorithm \ref{alg:ood_detection}  for GP-based OOD detection.

\begin{algorithm}
\caption{OOD Detection via GP Predictive Distribution}
\textbf{Input:} The test point $\mathbf{x}'$, the training data $X_{\rm{InD}}^k$, $\mathbf{z}_{\rm{InD}}^k$, $k=1, \ldots, K$, and the pre-specified true positive rate (TPR), $1-\alpha$.

\begin{enumerate}
    \item \textbf{Split the training data:} For each $X_{\rm{InD}}^k$, divide it into subsets for GP fitting and OOD detection rule, with sizes $m^{\rm{GP}}_k$ and $m^{\rm{Valid}}_k$ ($n_k = m^{\rm{GP}}_k + m^{\rm{Valid}}_k$), i.e., 
    \[
    X_{\rm{InD}}^k = X_{\rm{GP}}^k \cup X_{\rm{Valid}}^k.
    \]
    Denote the output at $X_{\rm{GP}}^k$ as 
    \[
    \mathbf{z}_{\rm{GP}}^k := \{f_k(\mathbf{x}): \mathbf{x} \in X_{\rm{GP}}^k\} \in \mathbb{R}^{m^{\rm{GP}}_k \times 1}.
    \]

    \item \textbf{Fit Gaussian Processes (GPs):} For each $k$, fit a GP using the input-output pairs $\left(\boldsymbol{\xi}(X_{\rm{GP}}^k), \mathbf{z}_{\rm{GP}}^k\right)$, where the hyper-parameters $\theta_{k,j}$ and $\tau^2_k$ are estimated via MLE. The resulting predictive distribution is a normal distribution $\mathcal{N}(\mu_k(\mathbf{x}), \sigma^2_k(\mathbf{x}))$, with:
    \begin{equation}\label{eq:GPmean}
        \mu_k(\mathbf{x}) = \Phi_k(\boldsymbol{\xi}(\mathbf{x}), \boldsymbol{\xi}(X_{\rm{GP}}^k)) \Phi_k(\boldsymbol{\xi}(X_{\rm{GP}}^k), \boldsymbol{\xi}(X_{\rm{GP}}^k))^{-1} \mathbf{z}_{\rm{GP}}^k,
    \end{equation}
    and
    \begin{equation}\label{eq:GPvariance}
        \sigma_k^2(\mathbf{x}) = \hat{\tau}^2_k\left[1 - \Phi_k(\boldsymbol{\xi}(\mathbf{x}), \boldsymbol{\xi}(X_{\rm{GP}}^k)) \Phi_k(\boldsymbol{\xi}(X_{\rm{GP}}^k), \boldsymbol{\xi}(X_{\rm{GP}}^k))^{-1} \Phi_k(\boldsymbol{\xi}(X_{\rm{GP}}^k), \boldsymbol{\xi}(\mathbf{x}))\right].
    \end{equation}

    \item \textbf{Determine the critical value $\gamma_k$:} For $k=1, \ldots, K$, compute:
    \[
    \gamma_k = q_\alpha\left(s_k(\mathbf{x}') : \mathbf{x}' \in X_{\text{Valid}}^k\right),
    \]
    where 
    \[
    s_k(\mathbf{x}') = \frac{1}{m^{\rm{Valid}}_k} \sum_{\mathbf{x} \in X_{\rm{Valid}}^k} d(\mathbf{x}', \mathbf{x}),
    \]
    and $d(\mathbf{x}', \mathbf{x})$ is defined as in Equation \eqref{eq:KL}.
\end{enumerate}
\textbf{Output:} For a new test point $\mathbf{x}'$ predicted to belong to class $k$, i.e., $k=\text{arg}\,\max_{l=1,\ldots,K}{f}_{l}(\mathbf{x}')$ the detection function is:
    \begin{equation*}
    g(\mathbf{x}') = 
    \begin{cases} 
    \rm{OOD}, & \text{if } s_k(\mathbf{x}') > \gamma_k, \\
    \rm{InD}, & \text{if } s_k(\mathbf{x}') \leq \gamma_k.
    \end{cases}
    \end{equation*}
    \label{alg:ood_detection}
\end{algorithm}

\subsection{Choice of the Intermediate Layer}\label{ss:intermediate}
Training the multi-class GP requires selecting a pre-specified intermediate layer $\boldsymbol{\xi}(\textbf{x})$ as the model input. Given the ultimate goal of separating InD and OOD data using the score function $s(\mathbf{x})$, we aim to find a transformation $\boldsymbol\xi$ such that the minimum distance between the sets $\left\{\boldsymbol\xi(\mathbf{x})|\mathbf{x}\in\mathcal{X}_\text{InD}\right\}$ and $\left\{\boldsymbol\xi(\mathbf{x})|\mathbf{x}\in\mathcal{X}_\text{OOD}\right\}$ is positive. This property is naturally satisfied when $\boldsymbol\xi(\mathbf{x})=\mathbf{x}$, as the objects (e.g., images) are inherently distinct in the original data space.

The minimum distance is formally defined as: 
\begin{equation}\label{eq:mindist}
    d_{\min,k}(\mathbf{x}')=\min_{\mathbf{x}\in X_{\text{InD}}^k}\{\|\Theta_k^{-1}(\boldsymbol{\xi}(\mathbf{x})-\boldsymbol{\xi}(\mathbf{x}'))\|_2\},
\end{equation}
where $\Theta_k={\rm{diag}}(\theta_{k,1},\ldots,\theta_{k,p})$.
The following theorem provides insights into the relationship between the minimum distance and the OOD detection function $g(\mathbf{x}')$. Additional notations are provided in Algorithm \ref{alg:ood_detection}.
\begin{theorem} Suppose $\mathbf{x}'\in\mathcal{X}$ and $k=\text{arg}\,max_{l=1,\ldots,K}{f}_{l}(\mathbf{x}')$. Consider the squared exponential kernel as defined in \eqref{eq:kernel function}, and assume that the hyper-parameters $\Theta_k$ and $\tau^2_k$ are fixed. Then, it follows that $g(\mathbf{x}')=1$ if 
\begin{align*}
d_{\min,k}(\mathbf{x}')^2>-\frac{1}{2}\log\left(\frac{2a_k\lambda_{\min}(\boldsymbol{\Phi}_k)}{m^{\rm{GP}}_k}\right)
\end{align*}
where 
\begin{align*}
    a_k=\gamma_k-\frac{1}{m^{\rm{Valid}}_k}\sum_{\mathbf{x}\in X_{\rm{Valid}}^k}\log[ 1 - \Phi_k(\boldsymbol{\xi}(\mathbf{x}),\boldsymbol{\xi}(X_{\rm{GP}}^k))\boldsymbol{\Phi}_k^{-1}\Phi_k(\boldsymbol{\xi}(X_{\rm{GP}}^k),\boldsymbol{\xi}(\mathbf{x}))],
\end{align*}
$\boldsymbol{\Phi}_k=\Phi_k(\boldsymbol{\xi}(X_{\rm{GP}}^k), \boldsymbol{\xi}(X_{\rm{GP}}^k))$, and $\lambda_{\min}(\boldsymbol{\Phi}_k)$ denotes the minimum eigenvalue of $\boldsymbol{\Phi}_k$.
\end{theorem}

This theorem indicates that a sample $\mathbf{x}'$ will be identified as OOD if its distance from the InD samples, measured as $\|\Theta_k^{-1}(\boldsymbol{\xi}(\mathbf{x})-\boldsymbol{\xi}(\mathbf{x}'))\|_2$, exceeds a threshold. The lower bound of this distance is influenced by $\lambda_{\min}(\boldsymbol{\Phi}_k)$, the smallest eigenvalue of the kernel matrix $\boldsymbol{\Phi}_k$, which relates to the space-fillingness of the input data. When $\lambda_{\min}(\boldsymbol{\Phi}_k)$ is large, the input points are well-distributed (space-filling), resulting in a kernel matrix with less redundancy (less collinearity). In this case, the lower bound on the minimum distance becomes smaller, meaning that a sample will be identified as OOD even if it is only slightly far from the InD samples. Conversely, when $\lambda_{\min}(\boldsymbol{\Phi}_k)$ is small, the input points are clustered or poorly distributed. In this scenario, the lower bound on the minimum distance becomes larger, requiring a sample to be significantly far from the InD samples to be classified as OOD. While this is not a strict ``if and only if'' condition, this relationship provides valuable insights into the separation distance between InD and OOD data. Nonetheless, while the theorem is logical, using the distance $\|\Theta_k^{-1}(\boldsymbol{\xi}(\mathbf{x})-\boldsymbol{\xi}(\mathbf{x}'))\|_2$  for separation between InD and OOD samples may not be optimal in practice. Potential improvements to this approach, incorporating more modern GP techniques, such as nonstationary GP models, which can adapt to varying data structures, are discussed in Section \ref{s:conclusion}.

This theorem also provides practical considerations for selecting $\boldsymbol\xi(\mathbf{x})$. When OOD samples $\mathbf{x}'\in\mathcal{X}_\text{OOD}$ are not available, the choice of $\boldsymbol\xi(\mathbf{x})$ must balance two aspects. On one hand, a lower-dimensional $\boldsymbol\xi(\mathbf{x})$ can improve the GP emulator's performance for predicting $\mathbf{f}$ but it may fail to adequately separate InD and OOD data. On the other hand, a higher-dimensional $\boldsymbol\xi(\mathbf{x})$ increases the minimum distance defined in \eqref{eq:mindist}, but it may also lead to higher computational costs for GP training and poorer GP fitting performance. We recommend to choose $\boldsymbol{\xi}(\mathbf{x})$ to have as high a dimensionality as possible, provided that the GP emulator achieves classification accuracy comparable to the original DNN. This ensures effective separation while maintaining reasonable computational costs. When OOD samples $\mathbf{x}'\in\mathcal{X}_\text{OOD}$ are available, data visualization techniques such as t-SNE \citep{van2008visualizing} or UMAP \citep{mcinnes2018umap} can be applied to the InD and OOD data to analyze their distribution in the feature space. This enables an informed choice of
$\boldsymbol{\xi}(\mathbf{x})$  that maximizes the separation between InD and OOD samples. Additional improvements for handling high-dimensional GPs are discussed in Section \ref{s:conclusion}.

\section{Experiments and Results}
\label{s:case}
We demonstrate the effectiveness of the proposed method under two distinct settings. In the first setting, we conduct tests on conventional image classification datasets to establish benchmarks and validate performance. In the second setting, we evaluate the method on large-scale, case-specific real-world image datasets \citep{tao2023non}. The OOD detection performance is evaluated based on the OOD detection accuracy for InD and OOD data as well as the area under receiver operating characteristic curve (AUROC). In both scenarios, the proposed approach successfully identifies OOD samples without exposure to OOD data during the training phase. The proposed method’s performance is also compared with several state-of-the-art OOD detection techniques, highlighting its robustness and competitive performance.

\subsection{Conventional Image Classification Datasets}
Since conventional image classification datasets do not involve OOD detection tasks, one dataset was selected as InD dataset for model training while the remaining datasets were considered as OOD test data in this setting. Specifically, the \textsf{MNIST} dataset \citep{lecun1998gradient} was considered as InD data, while \textsf{FashionMNIST} dataset \citep{xiao2017fashion}, \textsf{Cifar-10} \citep{krizhevsky2009learning}, and \textsf{SVHN} \citep{netzer2011reading}, and a downsampled version of ImageNet \citep{deng2009imagenet} were used as OOD datasets for test. In this experiment, the sample sizes in the training, validation, InD test and OOD test datasets are $10000$, $2000$, $3000$ and $3000$, respectively. 

\begin{figure*}
    \centering
    \includegraphics[scale=0.5]{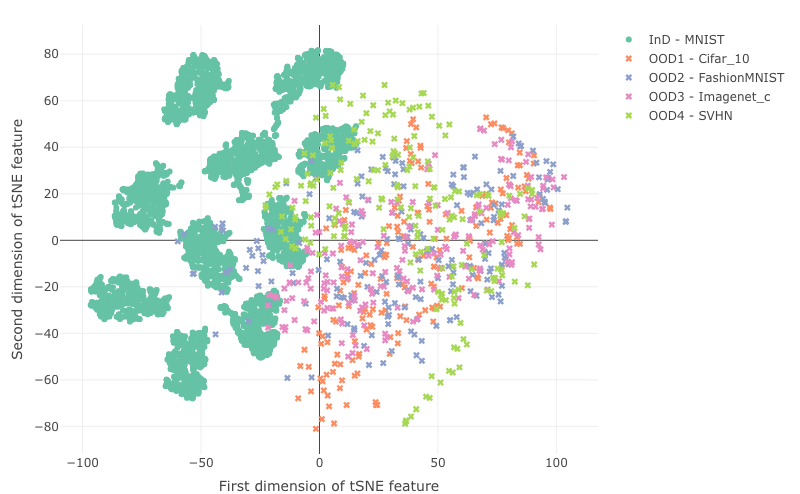}
    \captionsetup{font=small}
    \caption{Visualization of the InD and OOD data from the conventional image classification datasets using t-SNE. The InD data ($\circ$) forms 10 distinct clusters, corresponding to the 10 classes from the \textsf{MNIST} data. The OOD data ($\times$) are distinguishable from the InD data. Different colors represent different datasets.}
    \label{fig:Sep_MNIST}
\end{figure*}

A convolution DNN was first trained to classify images form the \textsf{MNIST} dataset, with a DenseNet architecture \cite{huang2017densely}. Following Subsection~\ref{ss:intermediate}, the intermediate layer of dimension $32$ was chosen to balance the computational efficiency and fitting accuracy of the multi-class GP. We further validated the dimension of the intermediate layer using UMAP visualization as shown in Figure~\ref{fig:Sep_MNIST}, where $\boldsymbol\xi$ of InD and OOD data were projected and visualized in a two-dimensional space. A clear separation between InD and OOD data is exhibited in Figure~\ref{fig:Sep_MNIST}. Notably, InD data forms 10 distinct clusters, which aligns with the 10 classes of \textsf{MNIST}. 

During training, only the \textsf{MNIST} dataset was used to optimize the model parameters in the DNN. Intermediate features and unconstrained Softmax scores from \textsf{MNIST} were saved after training and validation. The DNN were further used to extract the features $\boldsymbol\xi$ of the data from \textsf{FashionMNIST}, \textsf{Cifar-10}, \textsf{SVHN}, and \textsf{ImageNet} datasets.

For InD data, the extracted features served as inputs for training $K = 10$ GP models, while the unconstrained Softmax scores were treated as the GP responses. We utilized the \texttt{laGP} package \citep{gramacy2015local,gramacy2016lagp} in \texttt{R} \citep{R2023} for GP training with the \textsf{MNIST} dataset. A separate validation dataset was used to evaluate the prediction variance of GP at InD data points and hence compute the detection score $s$. The thresholds $\gamma_k$ were then determined as quantiles of the scores $s$ within each class $k$ to guarantee a pre-specified true positive rate (TPR) $1-\alpha$. In the inference stage, model parameters and the thresholds remained fixed, with the multi-class GP model providing predicted means and variances. The OOD detection scores were then calculated and compared with the threshold to make the final detection decision. All setups above ensure that OOD data was not involved in any training or validation process. 

The trained OOD detection function was then tested with multiple OOD datasets. The performance was evaluated according to TPR, True Negative Rate (TNR) and AUROC. The results are displayed in Table~\ref{tab:mnist_results}, with the true positive rate set as $90\%$ (the first row) and $95\%$ (the second row), respectively.
{\footnotesize
\begin{table}[htbp]
    \centering
    \caption{OOD detection performance in conventional image classification datasets.}
    \label{tab:mnist_results}
    \begin{tabular}{|c|c|c|c|c|}
        \hline
        TPR & \begin{tabular}{c}\textsf{FashionMNIST}\\TNR/AUROC\end{tabular} 
        & \begin{tabular}{c}\textsf{Cifar10}\\TNR/AUROC\end{tabular} 
        & \begin{tabular}{c}\textsf{SVHN}\\TNR/AUROC\end{tabular} 
        & \begin{tabular}{c}\textsf{ImageNet}\\TNR/AUROC\end{tabular} 
      \\
        \hline
        0.9036 & 0.8286/0.8661 & 1.0000/0.9518 & 1.0000/0.9518 & 1.0000/0.9518 \\
        \hline
        0.9546 & 0.7556/0.8551 & 1.0000/0.9773 & 1.0000/0.9773 & 0.9996/0.9771\\
        \hline
    \end{tabular}
\end{table}
}
Table~\ref{tab:mnist_results} indicates that the proposed method achieved high OOD detection accuracy across \textsf{Cifar10}, \textsf{SVHN} and \textsf{ImageNet}. The AUROC scores across these datasets remained relatively stable, around 0.93 at a $90\%$ TPR and over 0.95 at a $95\%$ TPR, suggesting a robust OOD detection performance at high TPR without seeing any OOD samples. The results also show that increasing the $1-\alpha$ from 0.9 to 0.95 improves both TPR (from 0.8795 to 0.9420) and AUROC across all OOD datasets. This indicates that a stricter threshold allows the model to better balance the trade-off between detecting OOD samples and maintaining InD accuracy for the image datasets. The performance on \textsf{FashionMNIST} as an OOD dataset is worse than other datasets in terms of AUROC, which could be attributed to its similarity to the InD dataset - \textsf{MNIST}. Nonetheless, the AUROC remains high, demonstrating that the proposed approach maintains reliable OOD detection of incoming OOD samples similar as the InD samples.

The proposed method was compared against several established OOD detection methods. All the methods were evaluated under the setup where OOD samples were not visible in the training stage. These methods include ODIN \citep{liang2018enhancing}, Mahalanobis Distance-based Detection \citep{lee2018simple}, Energy-Based OOD Detection \citep{liu2020energy}, and VOS (Virtual Outlier Synthesis, \cite{du2022vos}), the descriptions of which can be found in Section~\ref{sec:intro}. Note that the TNR and AUROC of ODIN and Mahalanobis Distance-based Detection were calculated as the average of all the TNR and AUROC using different hyper-parameters since tuning the hyper-parameters requires OOD data. We set the temperature parameter in \cite{liu2020energy} and \cite{du2022vos} as the nominal value because the two papers claim to be hyper-parameter free. We set TPR as $0.95$ in the training stage and compare the TNR and AUROC for OOD data. The comparison results are displayed in Table~\ref{tab:conventional_tnr_tpr95}.

%(i) ODIN \cite{liang2018enhancing} that improves OOD detection through temperature scaling and input perturbation applied to softmax scores, (ii) Mahalanobis Distance-based Detection \cite{lee2018simple} that models class-conditional Gaussian distributions of features using the Mahalanobis distance, (iii) Energy-Based OOD Detection \cite{liu2020energy} that distinguishes InD and OOD samples by analyzing energy scores from the model's output, with lower energy values indicating InD data, and (iv) VOS (Virtual Outlier Synthesis) \cite{du2022vos} that generates virtual OOD samples during training to enhance the model’s capacity to separate InD and OOD data. Multi-class GP is our proposed approach that leverages Gaussian processes to model the distribution of features and unnormalized softmax scores.

\begin{table}[htbp]
    \centering
    \caption{Comparison with existing methods in conventional image classification datasets.}
    \label{tab:conventional_tnr_tpr95}
    \resizebox{\textwidth}{!}{%
    \begin{tabular}{|l|c|c|c|c|}
        \hline
        \begin{tabular}{l}TNR/AUROC\end{tabular} 
        & \begin{tabular}{l}\textsf{FashionMNIST}\end{tabular} 
        & \begin{tabular}{l}\textsf{Cifar10}\end{tabular} 
        & \begin{tabular}{l}\textsf{SVHN}\end{tabular} 
        & \begin{tabular}{l}\textsf{ImageNet}\end{tabular} \\
        \hline
        ODIN & 0.0591/0.2500 & 0.0237/0.0951 & 0.0234/0.0907 & 0.0034/0.0422 \\
        \hline
        Mahalanobis & 0.3422/0.7017 & 0.0000/0.5000 & 0.0000/0.5000 & 0.0000/0.5000 \\
        \hline
        Energy & 0.6110/0.8157 & 0.9980/0.9992 & 0.9785/0.9947 & 0.9870/0.9963 \\
        \hline
        VOS & 0.5085/0.7182 & 1.0000/1.0000 & 0.9855/0.9966 & 0.9995/0.9999 \\
        \hline
        Multi-class GP & 0.7556/0.8551 & 1.0000/0.9773 & 1.0000/0.9773 & 0.9996/0.9771 \\
        \hline 
    \end{tabular}%
    }

\end{table}

% TPR 0.95, 0.95, 0.9410, 0.9465, 0.9545
All the existing methods return close to $0.95$ TPR at the test dataset of InD data. Regarding the OOD detection performance, the proposed multi-class GP method achieves the highest TNR compared to the other existing methods while maintaining a comparable AUROC to VOS in \textsf{Cifar10}, \textsf{SVHN} and \textsf{ImageNet}. Importantly, the proposed method significantly outperforms the other methods in detecting OOD images from the \textsf{FashionMNIST} dataset, where the OOD samples share more similarity with the InD data from the \textsf{MNIST} dataset. These results highlight the strong capability of the proposed method in effectively separating InD and various OOD data.

\subsection{Real-World Image Datasets}
The second setup aligns with the large-scale OOD learning studies, where \textsf{ImageNet} \citep{imagenet} was used as the InD dataset, while Describable Textures Dataset (\textsf{DTD}, \cite{dtd}), \textsf{iSUN} \citep{xu2015turkergaze}, \textsf{LSUN} \citep{lsun}, \textsf{Places365} \citep{places365}, and SVHN \citep{netzer2011reading} served as the OOD datasets. The sample sizes in the training, validation, InD test and OOD test datasets are $10000$, $1500$, $1600$ and $1600$, respectively. 

\begin{figure*}
    \centering
    \includegraphics[scale=0.5]{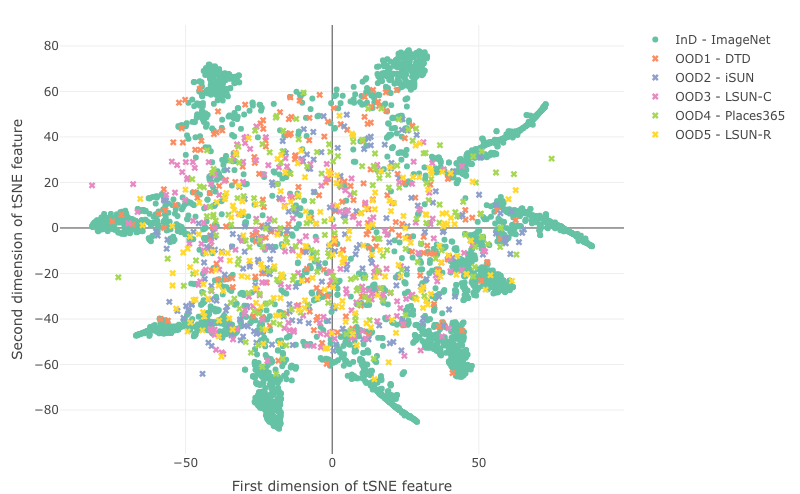}
    \captionsetup{font=small}
    \caption{Visualization of the InD and OOD data from the large-scale OOD learning studies using tSNE. The InD data ($\circ$) forms 10 distinct clusters, corresponding to the 10 classes from the \textsf{ImageNet} dataset. The majority of the OOD data ($\times$) are distinguishable from the InD data.}
    \label{fig:Sep_ImageNet}
\end{figure*}

All the images in the InD and OOD datasets were adjusted to have a shape of $3 \times 324 \times 324$ such that they can be classified using the same DNN. Detailed neural network architecture can be found in \citet{he2015deep}. The dimension of the intermediate layer $\boldsymbol\xi$ was set to $32$ as in the first experiment, indicating that the choice of the dimension of $\boldsymbol\xi$ does not depend on the DNN architecture nor the OOD data. $\boldsymbol\xi$ was further projected into the two dimensional space to visualize the separability via tSNE. The results are shown in Figure~\ref{fig:Sep_ImageNet} where slight overlap between InD and OOD classes are observed due to the data complexity.

Similar to the previous setup, the performance was assessed using TNR and AUROC by setting TPR to $90\%$ and $95\%$. The OOD detection results are shown in Table \ref{tab:imagenet_results}.

\begin{table}[htbp]
    \centering
    \caption{OOD detection performance in real-world image datasets.}
    \label{tab:imagenet_results}
    \resizebox{\textwidth}{!}{%
    \begin{tabular}{|c|c|c|c|c|c|c|}
        \hline
        TPR & \begin{tabular}{c}\textsf{DTD}\\TNR/AUROC\end{tabular} 
        & \begin{tabular}{c}\textsf{iSUN}\\TNR/AUROC\end{tabular} 
        & \begin{tabular}{c}\textsf{LSUN-C}\\TNR/AUROC\end{tabular} 
        & \begin{tabular}{c}\textsf{LSUN-R}\\TNR/AUROC\end{tabular} 
        & \begin{tabular}{c}\textsf{Places365}\\TNR/AUROC\end{tabular} 
        & \begin{tabular}{c}\textsf{SVHN}\\TNR/AUROC\end{tabular}\\
        \hline
        0.8956 & 0.6168/0.7562 & 0.5993/0.7446 & 0.6912/0.7906 & 0.6550/0.7725 & 0.5268/0.7084 & 0.2943/0.5921\\
        \hline
        0.9462 & 0.4168/0.6815 & 0.3412/0.6446 & 0.5112/0.7296 & 0.3687/0.6584 & 0.3062/0.6271 & 0.1781/0.5631 \\
        \hline
    \end{tabular}%
    }

\end{table}

At a 90$\%$ TPR, the proposed approach achieves close to $60\%$ TNR and $0.7$ AUROC across all the OOD datasets, with the best performance observed in \textsf{LSUN-C}. When the TPR is increased to 95$\%$, the average TNR drops to $40\%$ and AUROC decreased by $0.05$. The performance varies across different OOD datasets, with \textsf{LSUN-C} experiencing a large decline in TNR at the higher TPR. On the other hand, \textsf{iSUN} and \textsf{DTD} show moderate declines in TNR, suggesting that the two datasets are easier to be separated from InD. This can be validated in Figure~\ref{fig:Sep_ImageNet}, where \textsf{LSUN-C} has more overlapped area with \textsf{ImageNet}. Compared to the first setup, the worse OOD detection performance origins from the complex nature of the InD and OOD datasets - the inherent large variation within each class of the real-world objects brings about the difficulty in GP modeling and InD / OOD separation. 
% TPR 0.93, 0.95, 0.9431, 0.9556, 0.9500

The OOD detection performance was also compared among the proposed method and other existing methods. Table~\ref{tab:ood_tnr_tpr95} illustrates the comparison results. The proposed multi-class GP approach achieves comparable TNR and AUROC with ODIN among all the OOD datasets while showing advantages over Energy-Based OOD Detection and VOS. The Mahalanobis Distance-Based Detection method has the best detection performance because its decision threshold was determined through a logistic regression model that requires both InD and OOD labels. Nonetheless, neither ODIN nor Mahalanobis Distance-Based Detection method provide an over $0.5$ TNR in the first setup in Table~\ref{tab:conventional_tnr_tpr95}, showing that  the proposed method  achieves robust results overall.

\begin{table}[htbp]
    \centering
    \caption{Comparison with existing methods in real-world image datasets.}
    \label{tab:ood_tnr_tpr95}
    \resizebox{\textwidth}{!}{%
    \begin{tabular}{|l|c|c|c|c|c|c|}
        \hline
        TNR/AUROC & \begin{tabular}{l}\textsf{DTD}\end{tabular} 
        & \begin{tabular}{l}\textsf{iSUN}\end{tabular} 
        & \begin{tabular}{l}\textsf{LSUN-C}\end{tabular} 
        & \begin{tabular}{l}\textsf{LSUN-R}\end{tabular} 
        & \begin{tabular}{l}\textsf{Places365}\end{tabular} 
        & \begin{tabular}{l}\textsf{SVHN}\end{tabular}\\
        \hline
        ODIN & 0.4956/0.8017 &  0.5217/0.8163 & 0.5766/0.8100 & 0.5464/0.8296 &0.4125/0.7602 & 0.3095/0.6889 \\
        \hline
        Mahalanobis & 0.6100/0.9003 &  0.6763/0.9295 & 0.8739/0.9743 & 0.6755/0.9334 &0.1120/0.6455 & 0.6484/0.9383 \\
        \hline
        Energy & 0.0025/0.0935 & 0.0000/0.0794 & 0.0000/0.0492 & 0.0000/0.0679 &0.0000/0.0922 & 0.0000/0.0182 \\
        \hline
        VOS & 0.0081/0.1713 & 0.0012/0.1543 & 0.0000/0.0751 & 0.0000/0.1375 & 0.0000/0.1003 & 0.0000/0.0849\\
        \hline
        Multi-class GP &  0.4168/0.6815 & 0.3412/0.6446 & 0.5112/0.7296 & 0.3687/0.6584 & 0.3062/0.6271 & 0.1781/0.5631\\
        \hline
    \end{tabular}%
    }
\end{table}

\section{Conclusions and Discussions}\label{s:conclusion}

This paper introduces a novel approach for OOD detection using multi-class GP, which effectively emulates the DNN mapping between the intermediate layer and Softmax scores to achieve uncertainty quantification and detect OOD data based on the predictive distribution. Comprehensive experiments on conventional image classification datasets and large-scale real-world image datasets demonstrate that the proposed method outperforms several state-of-the-art OOD detection techniques without exposure to OOD data, particularly in terms of TNR and AUROC at  95$\%$ TPR. 

The experimental results highlight the advantages of probabilistic modeling inherent in multi-class GP, particularly in capturing the distribution of InD data and maintaining robustness across diverse OOD scenarios. This makes multi-class GP a versatile and reliable tool for applications where both accurate InD classification and robust OOD detection are essential.

Despite its promising results, research challenges remain. As discussed in Section \ref{ss:intermediate}, conventional GPs may struggle with high-dimensional and nonstationary data. Modern GP techniques, including deep GPs \citep{sauer2023vecchia,sauer2023active,ming2023deep}), Bayesian treed GPs \cite{gramacy2008bayesian}, active subspace GPs \citep{constantine2014active,garnett2014active,binois2022survey,binois2024combining}, additive GPs \citep{duvenaud2011additive,durrande2012additive}, clustered GPs \citep{Sung2019Clustered}, and multiple-output convolved GPs \citep{alvarez2011computationally}, offer potential improvements for modeling multi-class GPs by better capturing data heterogeneity. In addition, scalable approximations for large-scale datasets such as local GPs \citep{gramacy2015local}, Vecchia approximations \citep{katzfuss2021vecchia}, and global-local approximations \citep{vakayil2024global} can be explored to enhance scalability without compromising accuracy.

While this work focuses on OOD detection for image classification tasks, extending the method to other domains is a natural next step. Real-world applications, such as time-series analysis, healthcare, and autonomous systems, often involve OOD data with highly variable and unpredictable characteristics. We aim to address the remaining challenges of reducing sensitivity to adversarial prompts and further enhancing model reliability.

%It is also expected to extend the proposed approach to other tasks besides image classification, particularly in real-world scenarios where OOD data can be highly variable and unpredictable. We aim to address the remaining challenges of reducing sensitivity to adversarial prompts and further enhancing model reliability.

\label{sec:conc}
\appendix
\section{Appendix: Proof of theorems}
\textbf{Theorem 1 revisited:} Suppose $\mathbf{x}'\in\mathcal{X}$ and  $k=\text{arg}\,max_{l=1,\ldots,K}{f}_{l}(\mathbf{x}')$. Consider the squared exponential kernel as defined in \eqref{eq:kernel function}, and assume that the hyper-parameters $\Theta_k$ and $\tau^2_k$ are fixed. Then, it follows that $g(\mathbf{x}')=1$ if 
\begin{align*}
d_{\min,k}(\mathbf{x}')^2>-\frac{1}{2}\log\left(\frac{2a_k\lambda_{\min}(\boldsymbol{\Phi}_k)}{m^{\rm{GP}}_k}\right)
\end{align*}
where 
\begin{align*}
    a_k=\gamma_k-\frac{1}{m^{\rm{Valid}}_k}\sum_{\mathbf{x}\in X_\text{Valid}^k}\log[ 1 - \Phi_k(\boldsymbol{\xi}(\mathbf{x}),\boldsymbol{\xi}(X_{\rm{GP}}^k))\boldsymbol{\Phi}_k^{-1}\Phi_k(\boldsymbol{\xi}(X_{\rm{GP}}^k),\boldsymbol{\xi}(\mathbf{x}))],
\end{align*}
$\boldsymbol{\Phi}_k=\Phi_k(\boldsymbol{\xi}(X_{\rm{GP}}^k), \boldsymbol{\xi}(X_{\rm{GP}}^k))$, and $\lambda_{\min}(\boldsymbol{\Phi}_k)$ denotes the minimum eigenvalue of $\boldsymbol{\Phi}_k$.

\begin{proof} 
This proof follows the approach presented in \cite{sung2018exploiting}. 
The kernel function \eqref{eq:kernel function} can be written as $$\Phi_k(\boldsymbol{\xi}(\mathbf{x}),\boldsymbol{\xi}(\mathbf{x}'))=\phi(\|\Theta_k^{-1}(\boldsymbol{\xi}(\mathbf{x})-\boldsymbol{\xi}(\mathbf{x}'))\|_2),$$ where $\Theta_k={\rm{diag}}(\theta_{k1},\ldots,\theta_{kp})$, and $\phi(d)=\exp\{-d^2\}$. 

We start with the definition of the KL divergence in \eqref{eq:KL} where
\begin{equation}d\left(\mathbf{x}',\mathbf{x}\right)  = \log \frac{\sigma^2_k(\mathbf{x})}{\sigma^2_k(\mathbf{x}')} + \frac{\sigma^2_k(\mathbf{x}') + (\mu_k(\mathbf{x}') - \mu_k(\mathbf{x}))^2}{2 \sigma_k(\mathbf{x})^2} - \frac{1}{2},\nonumber
\end{equation}
and the OOD detection score defined as
\begin{equation}
    s_k(\mathbf{x}')=\frac{1}{m^{\rm{Valid}}_k}\sum_{\mathbf{x}\in X_\text{Valid}^k} d(\mathbf{x}',\mathbf{x}).
\end{equation}

If $s_k(\mathbf{x}')>\gamma_k$, it indicates that $\mathbf{x}'$ is  from OOD, which is equivalent to  
\begin{equation*}
  -\log \sigma^2_k(\mathbf{x}')+ \frac{1}{m^{\rm{Valid}}_k}\sum_{\mathbf{x}\in X_\text{Valid}^k} \frac{\sigma^2_k(\mathbf{x}') + (\mu_k(\mathbf{x}') - \mu_k(\mathbf{x}))^2}{2 \sigma_k(\mathbf{x})^2}> \gamma_k+ \frac{1}{2}-\frac{1}{m^{\rm{Valid}}_k}\sum_{\mathbf{x}\in X_\text{Valid}^k}\log\sigma^2_k(\mathbf{x}).
\end{equation*}

For notational simplicity, we denote that $\mathbf{w}=\boldsymbol{\xi}(\mathbf{x})$ and $W_{\rm{GP}}^k=\boldsymbol{\xi}(X_{\rm{GP}}^k)$ and $W_{\rm{Valid}}^k=\boldsymbol{\xi}(X_{\rm{Valid}}^k)$. Moreover, denote $\boldsymbol{\Phi}_k=\Phi_k(\boldsymbol{\xi}(X_{\rm{GP}}^k), \boldsymbol{\xi}(X_{\rm{GP}}^k))$.
 
From \eqref{eq:GPvariance}, we have
\begin{equation*}
\sigma_k^2(\mathbf{x}) =\tau^2_k[ 1 - \Phi_k(\mathbf{w},W_{\rm{GP}}^k)\boldsymbol{\Phi}_k^{-1}\Phi_k(W_{\rm{GP}}^k,\mathbf{w})]\leq \tau^2_k 
\end{equation*}
for any $\mathbf{x}\in X^k_{\rm{Valid}}$.
Thus, it follows that 
\begin{eqnarray}\label{eq:app1}
  &\quad&-\log \sigma^2_k(\mathbf{x}')+ \frac{1}{m^{\rm{Valid}}_k}\sum_{\mathbf{x}\in X_\text{Valid}^k} \frac{\sigma^2_k(\mathbf{x}') + (\mu_k(\mathbf{x}') - \mu_k(\mathbf{x}))^2}{2 \sigma_k(\mathbf{x})^2}\nonumber\\
  &>&-\log \tau^2_k+\frac{\sigma^2_k(\mathbf{x}') + \frac{1}{m^{\rm{Valid}}_k}\sum_{\mathbf{x}\in X_\text{Valid}^k} (\mu_k(\mathbf{x}') - \mu_k(\mathbf{x}))^2}{2 \tau^2_k}.
\end{eqnarray}

Since 
$$
\Phi_k(\mathbf{w}',W_{\rm{GP}}^k)\boldsymbol{\Phi}_k^{-1}\Phi_k(W_{\rm{GP}}^k,\mathbf{w}')\leq\|\Phi_k(\mathbf{w}',W_{\rm{GP}}^k)\|^2_2/\lambda_{\min}(\boldsymbol{\Phi}_k),
$$
where $\lambda_{\min}(\boldsymbol{\Phi}_k)$ is the minimum eigenvalue of $\boldsymbol{\Phi}_k$, we have
\begin{align}\label{eq:app2}
    \sigma^2_{k}(\mathbf{x}')&= \tau_k^2[1-\Phi_k(\mathbf{w}',W_{\rm{GP}}^k)\boldsymbol{\Phi}_k^{-1}\Phi_k(W_{\rm{GP}}^k,\mathbf{w}')]\nonumber\\
    &\geq \tau_k^2(1-\|\Phi_k(\mathbf{w}',W_{\rm{GP}}^k)\|^2_2/\lambda_{\min}(\boldsymbol{\Phi}_k)).
\end{align}

According to the definition $d_{\min,k}(\mathbf{x}')$ of the minimum (Mahalanobis-like) distance as in \eqref{eq:mindist}, it follows that  
$$
\Phi_k(\mathbf{w}',\mathbf{w})=\Phi_k(\boldsymbol\xi(\mathbf{x}), \boldsymbol\xi(\mathbf{x}'))\leq\phi(d_{\min,k}(\mathbf{x}')) \quad \text{for all}\quad \mathbf{x}\in X_{\rm{InD}}^k=X_{\rm{GP}}^k\cup X_{\rm{Valid}}^k,
$$
where  $\phi(d)=\exp\{-d^2\}$.
This implies that 
$$
\|\Phi_k(\mathbf{w}',W_{\rm{GP}}^k)\|^2_2\leq m^{\rm{GP}}_k\phi(d_{\min,k}(\mathbf{x}'))^2,
$$
which in turn gives that 
$$
\sigma^2_{k}(\mathbf{x}')\geq \tau_k^2\left[1-m^{\rm{GP}}_k\phi(d_{\min,k}(\mathbf{x}'))^2/\lambda_{\min}(\boldsymbol{\Phi}_k)\right].
$$

Since 
$$
\frac{\sum_{\mathbf{x}\in X_\text{Valid}^k} (\mu_k(\mathbf{x}') - \mu_k(\mathbf{x}))^2}{2 m^{\rm{Valid}}_k\tau^2_k}\geq 0,
$$
for any $\mathbf{x}$ and $\mathbf{x}'$, by combining \eqref{eq:app1} and \eqref{eq:app2},
we have 
\begin{align*}
&-\log \sigma^2_k(\mathbf{x}')+ \frac{1}{m^{\rm{Valid}}_k}\sum_{\mathbf{x}\in X_\text{Valid}^k} \frac{\sigma^2_k(\mathbf{x}') + (\mu_k(\mathbf{x}') - \mu_k(\mathbf{x}))^2}{2 \sigma_k(\mathbf{x})^2}\\
  >&-\log \tau^2_k+\frac{\sigma^2_k(\mathbf{x}') + \frac{1}{m^{\rm{Valid}}_k}\sum_{\mathbf{x}\in X_\text{Valid}^k} (\mu_k(\mathbf{x}') - \mu_k(\mathbf{x}))^2}{2 \tau^2_k}\\
  \geq& -\log \tau^2_k+\frac{1-m^{\rm{GP}}_k\phi(d_{\min,k}(\mathbf{x}'))^2/\lambda_{\min}(\boldsymbol{\Phi}_k)}{2}.
\end{align*}
Denote 
\begin{align*}
    a_k&=\log \tau^2_k+\gamma_k-\frac{1}{m^{\rm{Valid}}_k}\sum_{\mathbf{x}\in X_\text{Valid}^k}\log\sigma^2_k(\mathbf{x})\\
    &=\gamma_k-\frac{1}{m^{\rm{Valid}}_k}\sum_{\mathbf{w}\in W_\text{Valid}^k}\log[ 1 - \Phi_k(\mathbf{w},W_{\rm{GP}}^k)\boldsymbol{\Phi}_k^{-1}\Phi_k(W_{\rm{GP}}^k,\mathbf{w})],
\end{align*}
which is positive because $[ 1 - \Phi_k(\mathbf{w},W_{\rm{GP}}^k)\boldsymbol{\Phi}_k^{-1}\Phi_k(W_{\rm{GP}}^k,\mathbf{w})]< 1$ for any $\mathbf{w}$.
Then, since 
\begin{align*}
-\log \tau^2_k+\frac{1-m^{\rm{GP}}_k\phi(d_{\min,k}(\mathbf{x}'))^2/\lambda_{\min}(\boldsymbol{\Phi}_k)}{2}&> \gamma_k+ \frac{1}{2}-\frac{1}{m^{\rm{Valid}}_k}\sum_{\mathbf{x}\in X_\text{Valid}^k}\log\sigma^2_k(\mathbf{x}) \\
\Leftrightarrow \phi(d_{\min,k}(\mathbf{x}'))^2&< \frac{2a_k\lambda_{\min}(\boldsymbol{\Phi}_k)}{m^{\rm{GP}}_k}.
\end{align*}
Thus, we have that if $\phi(d_{\min,k}(\mathbf{x}'))^2< \frac{2a_k\lambda_{\min}(\boldsymbol{\Phi}_k)}{m^{\rm{GP}}_k}$,
which is equivalent to 
\begin{align*}
d_{\min,k}(\mathbf{x}')^2>-\frac{1}{2}\log\left(\frac{2a_k\lambda_{\min}(\boldsymbol{\Phi}_k)}{m^{\rm{GP}}_k}\right)
\end{align*}
because $\phi(d)=\exp\{-d^2\}$, which is a strictly decreasing function, it implies $s(\mathbf{x}')>\gamma_k$.
\end{proof}

\bibliographystyle{apalike}
\bibliography{Bibliography.bib}

\end{document}